\documentclass[runningheads]{llncs}
\usepackage{graphicx,subfigure}
\usepackage{comment}
\usepackage{amsmath,amssymb} 
\usepackage{color}
\usepackage[accsupp]{axessibility} 

\usepackage{booktabs}
\usepackage[pagebackref,breaklinks,colorlinks]{hyperref}
\usepackage[capitalize]{cleveref}

\definecolor{todocolor}{RGB}{255,0,0}

\definecolor{abcolor}{RGB}{0,0,255}

\begin{document}
\pagestyle{headings}
\mainmatter
\def\ECCVSubNumber{1336} 

\title{RC-MVSNet: Unsupervised Multi-View Stereo with Neural Rendering} % 
%******************

\titlerunning{RC-MVSNet}
\author{Di Chang\inst{1}, 
Aljaž Božič\inst{1},
Tong Zhang\inst{2},  
Qingsong Yan\inst{3},
Yingcong Chen\inst{3}, 
Sabine Süsstrunk\inst{2}, 
Matthias Nie{\ss}ner\inst{1}
}

\authorrunning{D. Chang et al.}
\institute{Technical University of Munich \and
École Polytechnique Fédérale de Lausanne
\and Hong Kong University of Science and Technology\\
        {\tt\small di.chang@tum.de}\quad
}

\maketitle

\begin{abstract}
Finding accurate correspondences among different views is the Achilles' heel of unsupervised Multi-View Stereo (MVS). Existing methods are built upon the assumption that corresponding pixels share similar photometric features. However, multi-view images in real scenarios observe non-Lambertian surfaces and experience occlusions. In this work, we propose a novel approach with neural rendering (RC-MVSNet) to solve such ambiguity issues of correspondences among views. Specifically, we impose a depth rendering consistency loss to constrain the geometry features close to the object surface to alleviate occlusions. Concurrently, we introduce a reference view synthesis loss to generate consistent supervision, even for non-Lambertian surfaces. Extensive experiments on DTU and Tanks\&Temples benchmarks demonstrate that our RC-MVSNet approach achieves state-of-the-art performance over unsupervised MVS frameworks and competitive performance to many supervised methods. The code is released at \url{https://github.com/Boese0601/RC-MVSNet}.

\end{abstract}

\begin{figure}
  \centering
    \subfigure[Unsup\_MVS~\cite{khot2019learning}]{
        \includegraphics[width=0.31\textwidth]{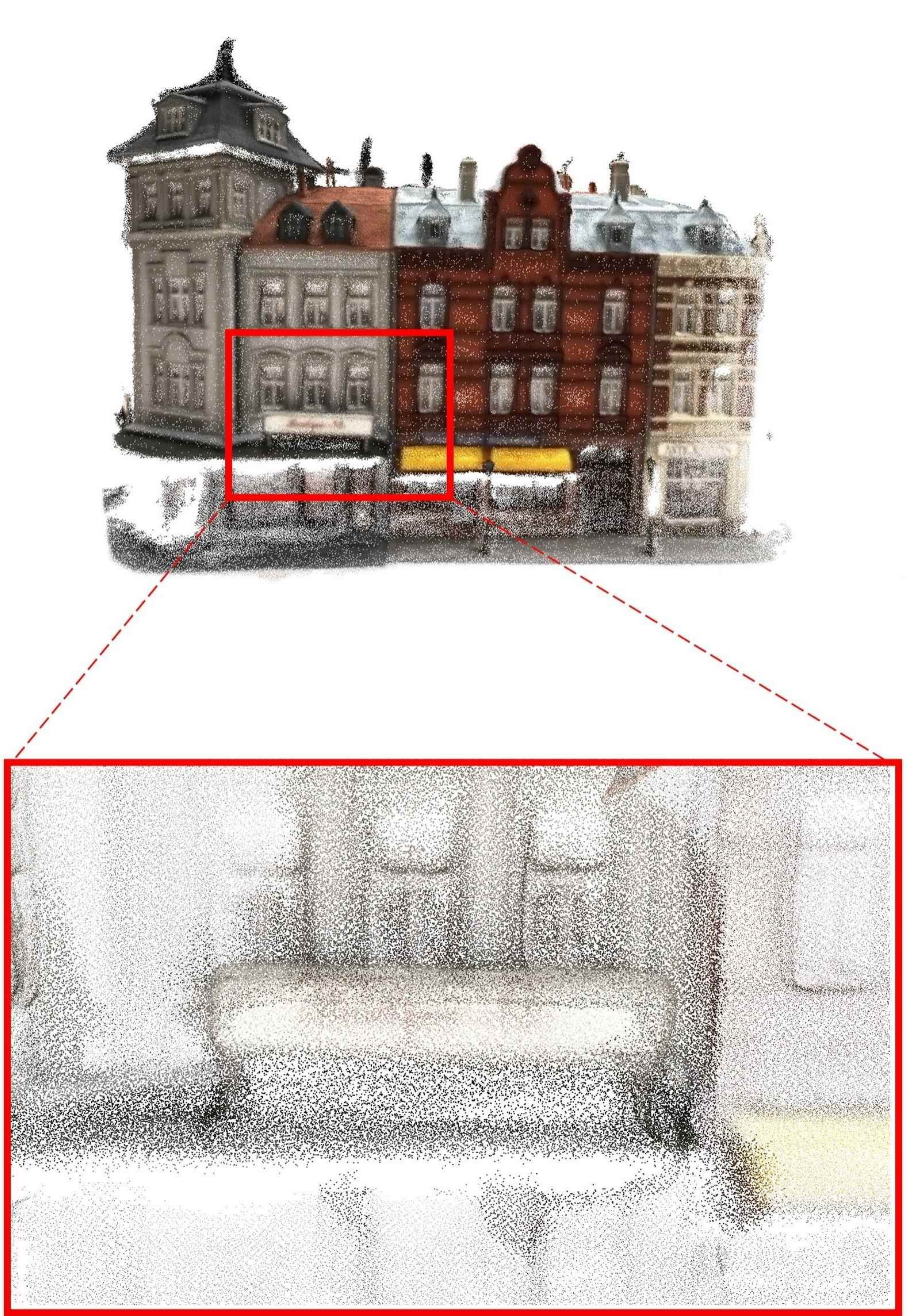}}
    \subfigure[JDACS~\cite{xu2021self}]{              
        \includegraphics[width=0.31\textwidth]{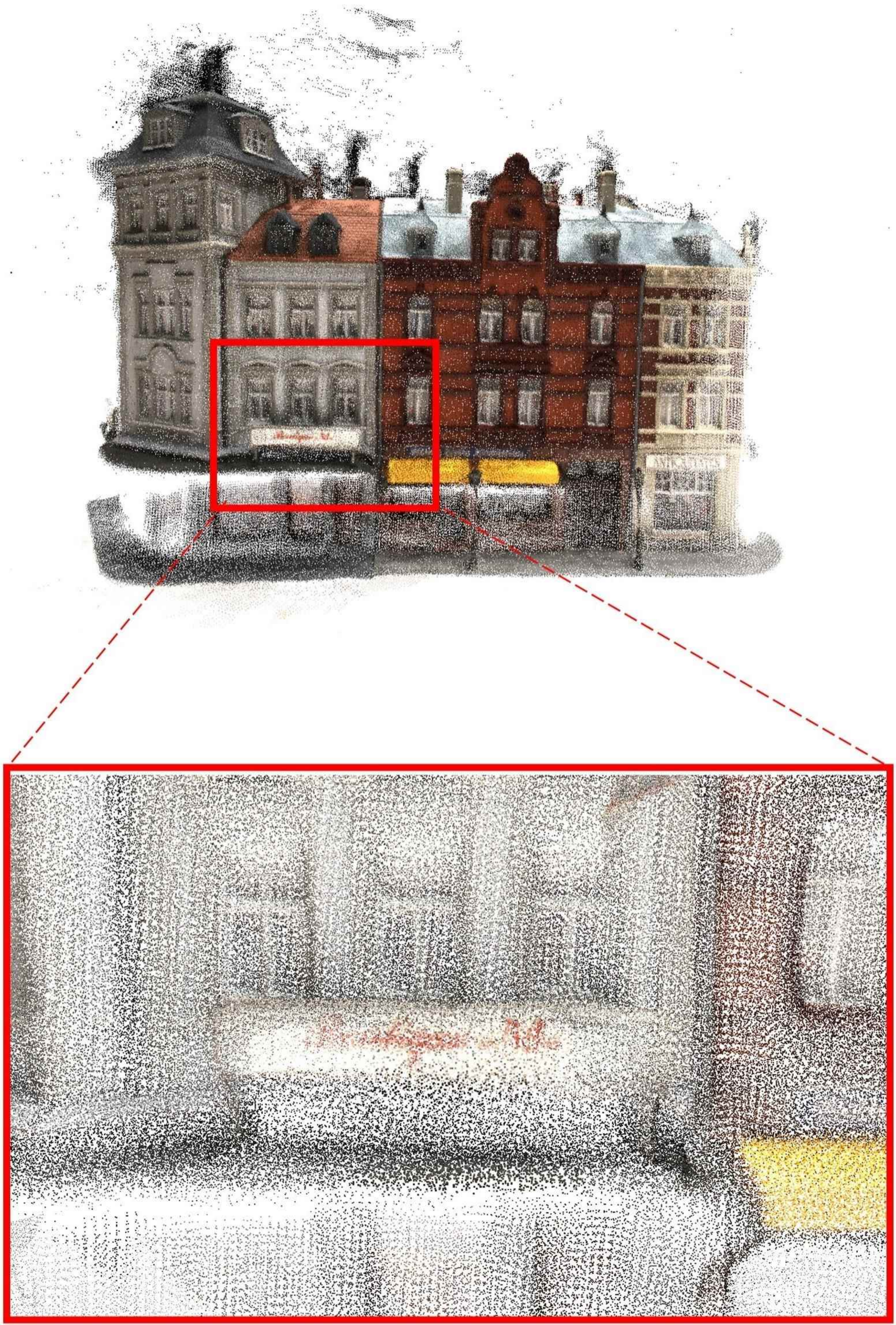}}
    \subfigure[Ours]{              
        \includegraphics[width=0.31\textwidth]{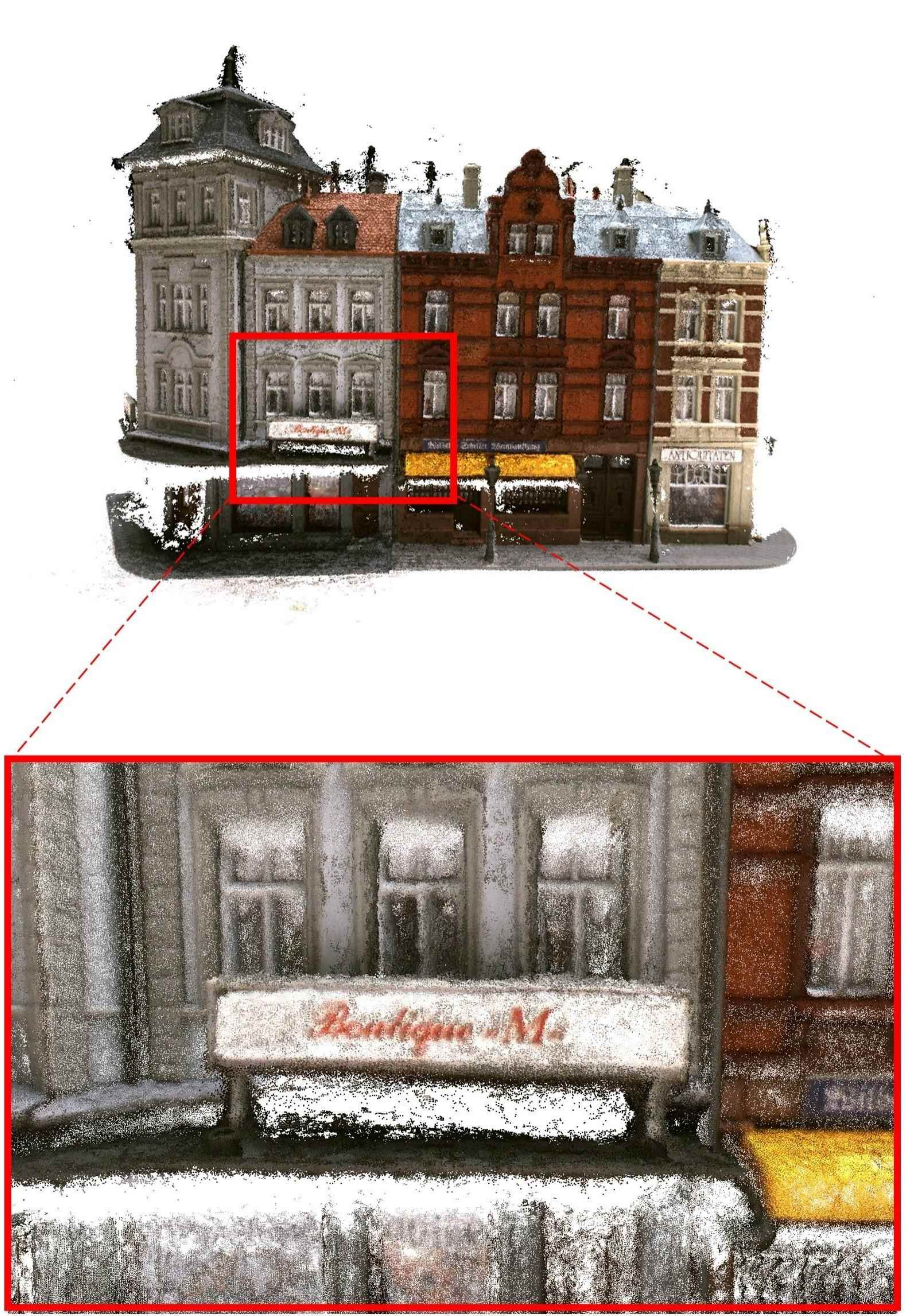}}

  \caption{Qualitative comparison of 3D reconstruction of our RC-MVSNet and previous unsupervised methods.~\cite{khot2019learning,xu2021self}}
     \label{fig:dtupoint}
\end{figure}

\section{Introduction}
\label{sec:intro}

Multi-View Stereo (MVS)~\cite{seitz2006comparison} is a long-standing and fundamental task in 3D computer vision. MVS aims to recover a 3D point cloud of real scenes from multi-view images and corresponding calibrated cameras. The widespread application of deep learning in recent years has lead to the emergence of end-to-end MVS depth estimation networks. A popular learning-based pipeline MVSNet~\cite{yao2018mvsnet} proposes to encode RGB information from different camera views into a cost volume and predict depth maps for point cloud reconstruction. Follow-up fully-supervised methods~\cite{gu2020cascade,mi2021generalized,wang2021patchmatchnet,xu2020pvsnet} further improved the neural network architecture, lowering memory usage and achieving state-of-the-art depth estimation performance on several benchmarks~\cite{aanaes2016large,knapitsch2017tanks}. However, these methods heavily rely on ground truth depth for supervision, which requires a depth sensor to collect the training data. It restricts these methods to limited datasets and largely indoor settings. To make MVS practical in more general real-world scenarios, it is vital to consider alternative unsupervised learning-based methods that can provide competitive accuracy compared to the supervised ones, while not requiring any ground truth depth.

\begin{figure}
    \centering
  \includegraphics[width=\textwidth]{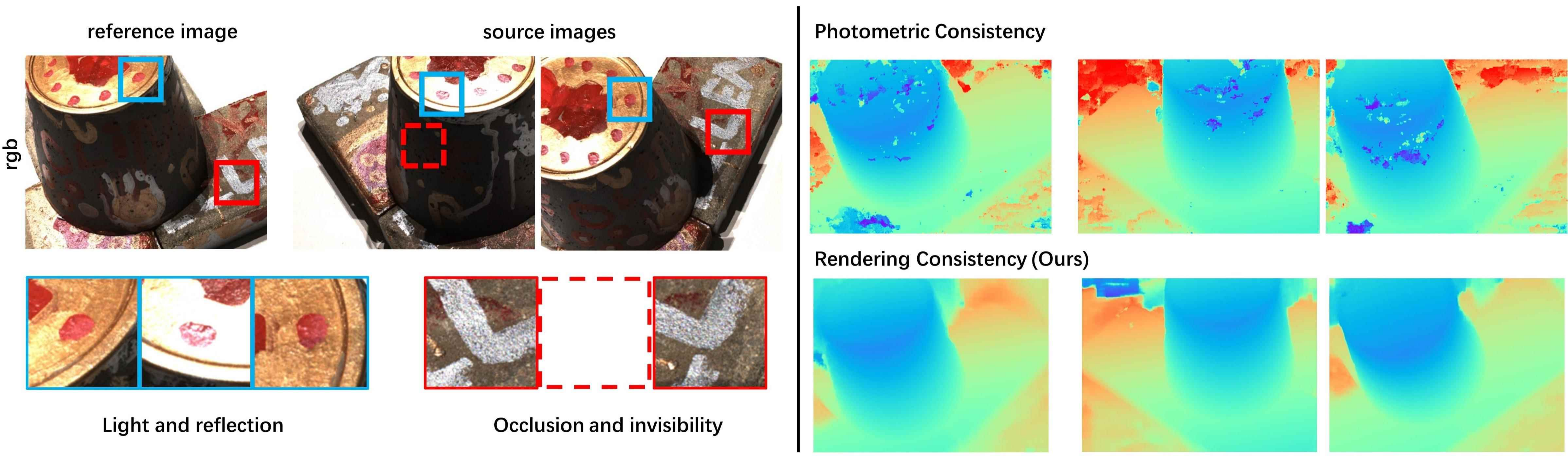}
  \caption{(left) Real-world photometric conditions and occlusions. (right) Top: Failure cases of a photometric consistency assumption. Bottom: The results of our proposed rendering consistency approach.}
  \label{fig:motivation}
\end{figure}
Existing unsupervised MVS methods~\cite{dai2019mvs2,huang2021m3vsnet,khot2019learning,xu2021self} are based on a \textit{photometric consistency hypothesis}, which states that pixels belonging to the same 3D point have identical color properties under different view directions. However, in a real-world environment, as shown in Fig.~\ref{fig:motivation},  occlusions, reflecting, non-Lambertian surfaces, varying camera exposure, and other variables will 
make such a hypothesis invalid. Thus, inaccurately matched points lead to false geometric reconstruction. Although Xu et al.~\cite{xu2021self} proposes to use semantic clues to remove the ambiguity in stereo matching, the improvement is very marginal due to its high dependency on the accuracy of a pretrained semantic features extraction model (with supervision). To remove inconsistencies with supervision, in this paper, we propose to use neural rendering to solve ambiguity in the case of view-dependent photometric effects and occlusions.

Recently, there has been an increasing interest in novel view synthesis, particularly with the introduction of Neural Radiance Fields~\cite{mildenhall2020nerf} that can model view-dependent photometric effects based on differentiable volumetric rendering. While initially focused on per-scene optimization with densely sampled input views, follow-up approaches~\cite{chen2021mvsnerf,rosu2021neuralmvs,wang2021ibrnet} propose to use a 2D CNN encoder that can be trained to predict novel views even with very few input images. Aside from the view-dependent radiance that defines the color along the viewing ray, these methods also learn a volume density, which can be interpreted as depth when integrated over the viewing ray. It is worth noting that the depth is learned in a purely unsupervised fashion. However, since the main goal is novel view synthesis, the depth obtained from volume density is often inaccurate. In our method, we build from the volumetric rendering approach proposed in~\cite{mildenhall2020nerf} with novel loss functions that can resolve view-dependent effects and occlusions, while keeping the depth representation used in MVS approaches to ensure locally accurate and smooth depth predictions.

To this end, we introduce RC-MVSNet, a novel end-to-end differentiable network for \textit{unsupervised multi-view stereo.} Combining the benefits of view-dependent rendering and structured depth representation, our method achieves state-of-the-art depth prediction results in the competitive DTU benchmark~\cite{aanaes2016large}, and also demonstrates robust performance on out-of-distribution samples in the Tanks and Temples dataset~\cite{knapitsch2017tanks}. In summary, our contributions are the following:
\begin{itemize}
    \item We propose a reference view synthesis loss based on neural volumetric rendering to generate RGB supervision that is able to account for view-dependent photometric effects. 

    \item We introduce Gaussian-Uniform mixture sampling to learn the geometry features close to the object surface to overcome occlusion artefacts present in existing unsupervised MVS approaches.
     \item We introduce a depth rendering consistency loss to refine the initial depth map by depth priors and ensure the robustness and smoothness of the prediction.
\end{itemize}

\section{Related Work}

\subsection{Supervised Multi-View Stereo}
Many \textit{supervised} approaches have proposed to use CNNs to interpret 3D scenes by predicting depth maps of the RGB inputs and reconstructing the point cloud using depth map filtering. Most state-of-the-art methods employ 3D cost volumes. As a representative work, MVSNet~\cite{yao2018mvsnet} encodes camera parameters and features into cost volumes through homography warping and regularizes the volume by 3D CNNs to generate depth prediction. Following works, e.g.,~\cite{gu2020cascade,wang2021patchmatchnet,yang2020cost} improve the performance of MVSNet and reduce the memory cost by introducing a multi-stage architecture and learn depth prediction in a coarse-to-fine manner. Furthermore, ~\cite{Wei_2021_ICCV,yan2020dense,yao2019recurrent} replace dense 3D convolution with convolutional recurrent GRU or LSTM units. However, the reliance on ground truth depth limits their application to specific datasets as discussed in Sec.~\ref{sec:intro}. Therefore, it is necessary to explore alternative unsupervised methods.

\subsection{Unsupervised and Self-supervised Multi-View Stereo}

End-to-end unsupervised and pseudo-label-based multi-stage self-supervised learning play pivotal roles in 3D vision, especially in multi-view reconstruction systems. The fundamental assumption of photometric consistency provides feasibility for unsupervised MVS. 
For instance, Unsup\_MVS~\cite{khot2019learning} proposed the first end-to-end learning-based unsupervised MVS framework: the images of source views are inversely warped to the reference view with its predicted depth map, and photometric consistency and SSIM~\cite{wang2004image} are enforced to minimize the discrepancy between the reference image and warped image. JDACS~\cite{xu2021self} proposes cross-view consistency of extracted semantic features and provides supervision of the segmentation map by non-negative matrix factorization. However, it still requires a pretrained semantic feature extraction backbone, and suffers from unstable convergence of cross-view semantic consistency that fails to provide reliable supervision.  U-MVSNet~\cite{xu2021digging} employs a flow-depth consistency loss to resolve ambiguous supervision. The dense 2D optical flow correspondences are used to generate pseudo labels for uncertainty-aware consistency, which improves the supervision to some extent. However, this method cannot be trained in an end-to-end  fashion, as it requires complex pre-training and fine-tuning. Self-supervised CVP-MVSNet~\cite{yang2021self} also proposes to learn depth pseudo labels with unsupervised pre-training, followed with the iterative self-supervised training to refine the pseudo label; however, it is still affected by ambiguous supervision. Furthermore, these unsupervised and self-supervised methods lack an occlusion-aware module to learn features from different viewing directions, leading to incomplete reconstruction of point clouds.

In contrast, our simple but effective model directly learns the geometric features of the scene by a reference view synthesis loss. This significantly reduces the training complexity and alleviates ambiguous photometric supervision. It also avoids the problems caused by occlusions through NeRF-like rendering.

\subsection{Multi-View Neural Rendering}
Recently, various neural rendering methods~\cite{bi2020deep,lombardi2019neural,mildenhall2020nerf,thies2019deferred,zhou2018stereo} have been presented, focusing on task of novel view synthesis. In particular, Neural Radiance Fields~\cite{mildenhall2020nerf} represent scenes with a continuous implicit function of positions and orientations for high quality view synthesis; several following up works~\cite{rebain2021derf,yu2021plenoxels,yu2021pixelnerf,zhang2020nerf++} improve both its efficiency and performance. There are also a few extensions~\cite{chen2021mvsnerf,rosu2021neuralmvs,wei2021nerfingmvs} of NeRF which introduce multi-view information to enhance the generalization ability of view synthesis. MVSNeRF~\cite{chen2021mvsnerf} proposes to leverage plane-swept cost volumes, which have been widely used in multi-view stereo, for geometry-aware scene understanding, and combines this with volumetric rendering for neural radiance field reconstruction; however, it fails to generate high-quality depth prediction in an unsupervised manner. NerfingMVS~\cite{wei2021nerfingmvs} uses sparse depth points from SfM for learning depth completion to guide the optimization process of NeRF. Our method takes advantage of both precise neural rendering and strong generalization of cost volumes and provides accurate depth estimation based on end-to-end unsupervised learning, which surpasses all previous unsupervised multi-view-stereo approaches and demonstrates accurate reconstruction of both indoor objects and out-of-distribution outdoor scenes.
 
\begin{figure}[t]
  \centering
  \includegraphics[width=\textwidth]{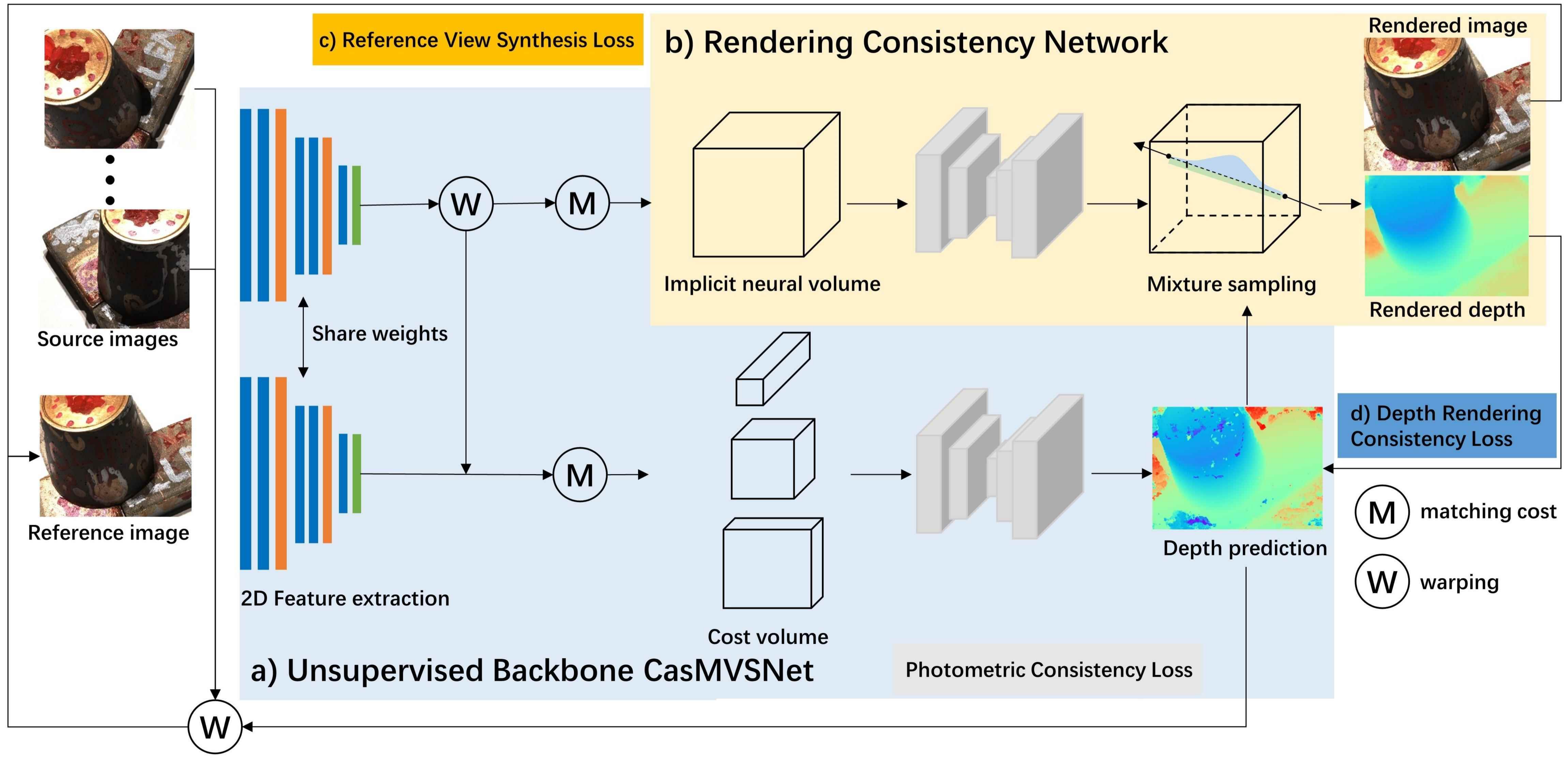}
  \caption{Overview of our unsupervised MVS approach (RC-MVSNet). a) Unsupervised Backbone CasMVSNet predicts initial depth map by photometric consistency and provides depth priors for rendering consistency network. b) Rendering Consistency Network generates image and depth by neural rendering under the guidance of depth priors. c) The rendered image is supervised by the reference view synthesis loss. d) The rendered depth is supervised by the depth rendering consistency loss.}
  \label{fig:pipeline}
\end{figure}

\section{Method}
\label{sec:method}

In this section, we describe RC-MVSNet, our proposed method for unsupervised multi-view stereo. Given $N$ images as input, with their corresponding cameras' intrinsic and extrinsic parameters, our method predicts a depth map in the reference camera view. The overall pipeline is illustrated in Fig.~\ref{fig:pipeline}. It consists of a backbone branch and an auxiliary branch. The backbone is built upon CasMVSNet~\cite{gu2020cascade}, which predicts a depth map in a coarse-to-fine manner, as described in Sec.~\ref{sec:photometric}. The auxiliary branch is built upon the neural radiance fields, which aside from geometry also models view-dependent scene appearance. Using volumetric rendering, the reference view is synthesized and compared to the input image (in Sec.~\ref{sec:reference}), resulting in accurate supervision even in the case of view-dependent effects. To ensure the geometric consistency between the network branches, an additional depth rendering consistency loss is introduced in Sec.~\ref{sec:depthrendering}, leading to more complete and accurate depth predictions, even in the case of occlusions. Note that during training, these two branches are simultaneously optimized, providing supervision to each other. During inference, only the backbone is used to obtain the depth prediction. We present the network optimization in detail in Sec.~\ref{sec:networkoptimization}.

\subsection{Unsupervised Multi-view Stereo Network}
\label{sec:photometric}

The backbone network closely follows the CasMVSNet~\cite{gu2020cascade} architecture. Input images $\left\{{I}_{j}\right\}_{j=1}^{N}$ are initially encoded with a shared 2D U-Net that generates pixel-wise features. Afterwards, a feature cost volume is constructed in the reference camera frustum.
Each volume voxel's position is projected into every input image using cameras' intrinsic and extrinsic parameters, and pixel features are queried  via bilinear interpolation. 
This results in warped feature volumes $\left\{{V}_{j}\right\}_{j=1}^{N}$ that are fused across views into a common feature volume $C$ by computing feature variance:
\begin{equation}
\label{eq:cost-volume}
{C}={Var}\left({V}_{1}, \cdots, {V}_{N}\right)=\frac{\sum_{j=1}^{N}\left({V}_{j}-\overline{{V}}_{j}\right)^{2}}{N}.
\end{equation}
The feature cost volume $C$ is further refined with a 3D U-Net and finally outputs a depth map in the reference view in a coarse-to-fine manner. Further details can be found in the supplementary material.

\subsubsection{Photometric Consistency.}
To supervise depth map prediction without any ground truth depth, existing methods~\cite{dai2019mvs2,huang2021m3vsnet,khot2019learning,xu2021digging,yang2021self} enforce photometric consistency between reference and other source input views. The key idea is to maximize the similarity between the reference image $I_{1}$ and any source image $ I_{j}$ after being warped to the reference view. Given corresponding intrinsics $K$, the relative transformation from reference to source view $T$, and predicted depth map $D$ in the reference view, we warp any pixel $p_i$ in the reference image (in a homogeneous notation) to the source image via inverse warping:
\begin{equation}
\hat{p_{i}}=K T\left(D(p_{i}) \cdot K^{-1} p_{i}\right).
\end{equation}
The warped source image $\hat{I}_{1}^{j}$ is synthesized by bilinearly sampling the source image's color values at the warped pixel location $\hat{p_{i}}$, i.e. $\hat{I}_{1}^{j}(p_{i})=I_{j}(\hat{p_{i}})$. In additon to the warped image $\hat{I}_{1}^{j}$, a binary mask $M_{j}$ is also generated, masking out invalid pixels that are projected outside the source image bounds. In the unsupervised MVS system, all $N-1$ source views are warped into the reference view to compute the photometric consistency loss:
\begin{equation}
\mathcal L_{P C}=\sum_{j=2}^{N} \frac{1}{\left\|M_{j}\right\|_{1}} \left(\left\|\left(\hat{I}_{1}^{j}-I_{1}\right) \odot M_{j}\right\|_{2}+\left\|\left(\nabla \hat{I}_{1}^{j}-\nabla I_{1}\right) \odot M_{j}\right\|_{2}\right).
\end{equation}
Here, $\nabla$ refers to a pixel-wise gradient and $\odot$ represents pixel-wise product.

\subsection{Reference View Synthesis}
\label{sec:reference}

Photometric consistency assumes consistent pixel colors across different input views, which does not hold in the case of view-dependent effects that are common with reflective materials. Furthermore, if certain pixels in the reference view are occluded in the matched source view, incorrect color similarity is enforced. To cope with these shortcomings, instead of using error-prone image warping of source images, we propose to learn the synthesis of reference views from source images using neural radiance fields~\cite{mildenhall2020nerf}, with the network learning how to take care of view-dependent effects and occlusions.

\subsubsection{Implicit Neural Volume Construction.}
Similar to the cost volume $C$ for the backbone branch, an implicit neural volume ${C^{\prime}}$ is also constructed via computation of feature variance, as defined in Eq.~\eqref{eq:cost-volume}. However, since we want to synthesize the reference view image using \textit{only} the information from the source views, we calculate the variance volume ${C^{\prime}}$ from warped volumes $\left\{{V}_{j}\right\}_{j=2}^{N}$, without the reference feature volume $V_{1}$. We thus gather features across $N-1$ source views, and apply a 3D U-Net $U(\cdot)$ to learn an implicit neural volume $F$ that we denote as $F=U(C^{\prime})$.

\subsubsection{Point Feature Aggregation.} 

To restore the scene geometry and appearance, rays are emitted from the reference camera center in the viewing direction of $\phi$. For each point $q \in \mathbb{R}^3$ along the ray we query RGB values $\ell = \{{I}_{j}[x_j,y_j]\}_{j=2}^{N}$ by bilinearly sampling source images,  where $x_j$ and $y_j$ represent the 2D location of point $q$, when it is projected to the image coordinate system of $j^{th}$ view. In addition to these properties, we also obtain a neural feature $f=F(q)$ from the implicit neural volume $F$ by trilinear interpolation at the location $q$. These neural features overcome the poor generalization of traditional neural radiance field approaches that require per-scene training or fine-tuning, by regressing properties from the geometric-aware cost volume (e.g. implicit neural volume in our case). This improves our method's robustness on datasets that are very different from the training set. These input features are concatenated into point features and we use an MLP $M(\cdot)$ to convert these features into predicted volume density $\sigma$ and color ${c}=[r, g, b]$. The whole process can be summarized as:
\begin{equation}
\sigma, c=M(q, \phi, f, \ell).
\end{equation}

Following previous works~\cite{chen2021mvsnerf,mildenhall2020nerf,roessle2021dense}, we encode the location $q$ and direction $d$ using positional encoding.

\subsubsection{Reference View Synthesis Loss.}

As in NeRF~\cite{mildenhall2020nerf}, we sample $K$ point candidates along the camera ray $\mathbf{r}(t)=\mathbf{o}+t\mathbf{\phi}$, where $\mathbf{o}$ origin is the camera and $\mathbf{\phi}$ is the pixel's viewing direction. For the sample distance $t_{k}$ in the range from near to far plane ($t_{n} \leq t_{k} \leq t_{f}$), we query the corresponding density $\sigma_k$ and color $c_k$. To compute pixel-wise color $\hat{\mathbf{C}}(\mathbf{r})$, color values are integrated along the ray $\mathbf{r}(t)$:
\begin{equation}
\label{eq:render}
\hat{\mathbf{C}}(\mathbf{r})=\sum_{k=1}^{K} w_{k} c_{k}.
\end{equation}
Here $ w_{k}=T_{k}\left(1-\exp \left(-\sigma_{k} \delta_{k}\right)\right)$ is the integration weight, where $\delta_{k}=t_{k+1}-t_{k}$ refers to the interval between adjacent sampling locations and $T_k$ indicates the accumulation of transmittance from $t_{n}$ to $t_{f}$:
\begin{equation}
\label{eq:transmittance}
T_{k}=\exp \left(-\sum_{k^{\prime}=1}^{k} \sigma_{k^{\prime}} \delta_{k^{\prime}}\right).
\end{equation}

To optimize the radiance field color output $\hat{\mathbf{C}}(\mathbf{r})$ in Eq.~\eqref{eq:render}, i.e. the synthesized RGB value of a pixel $(x, y)$ in the reference view, the rendering consistency loss $L_{RC}$ composed of a mean squared error is applied:
\begin{equation}
\mathcal L_{RC}=\|\hat{\mathbf{C}}(\mathbf{r})-\mathbf{C}(\mathbf{r})\|_{2}^{2}.
\end{equation}
Here ${\mathbf{C}}(\mathbf{r})$ refers to pixel color in the reference image ${I_{1}[x,y]}$.

\subsection{Depth Rendering Consistency}

Using an auxiliary rendering branch and applying rendering consistency loss significantly improves the quality of unsupervised depth predictions, as can be seen in our ablations in Sec.~\ref{sec.ablation}.
This results from joint training of both branches and optimizing a shared 2D image encoder. To further benefit  from rendering consistency and additionally improve unsupervised depth prediction, we also apply Gaussian-Uniform mixture sampling and a depth rendering consistency loss, which increase the beneficial coupling of both network branches.

\subsubsection{Gaussian-Uniform Mixture Sampling.}
\label{sec:depthrendering}
\begin{figure}

  \centering
    \subfigure[Uniform and Gaussian sampling]{
        \includegraphics[width=0.6\textwidth]{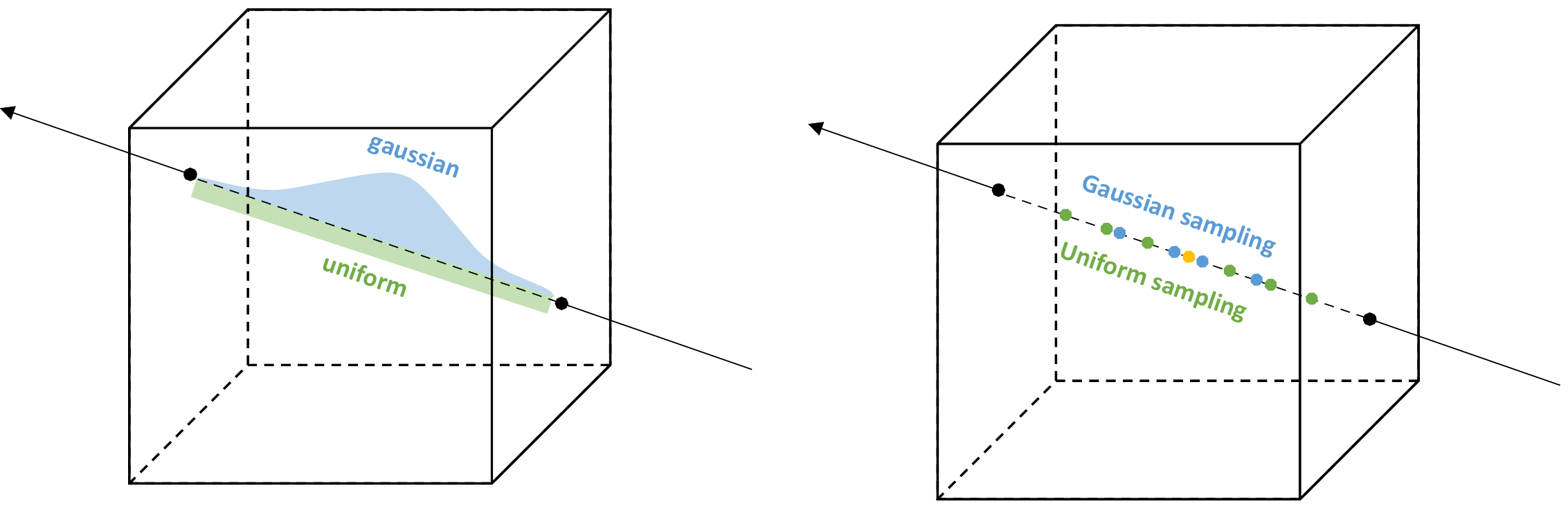}}
    \subfigure[Point Features Aggregation]{              
        \includegraphics[width=0.35\textwidth]{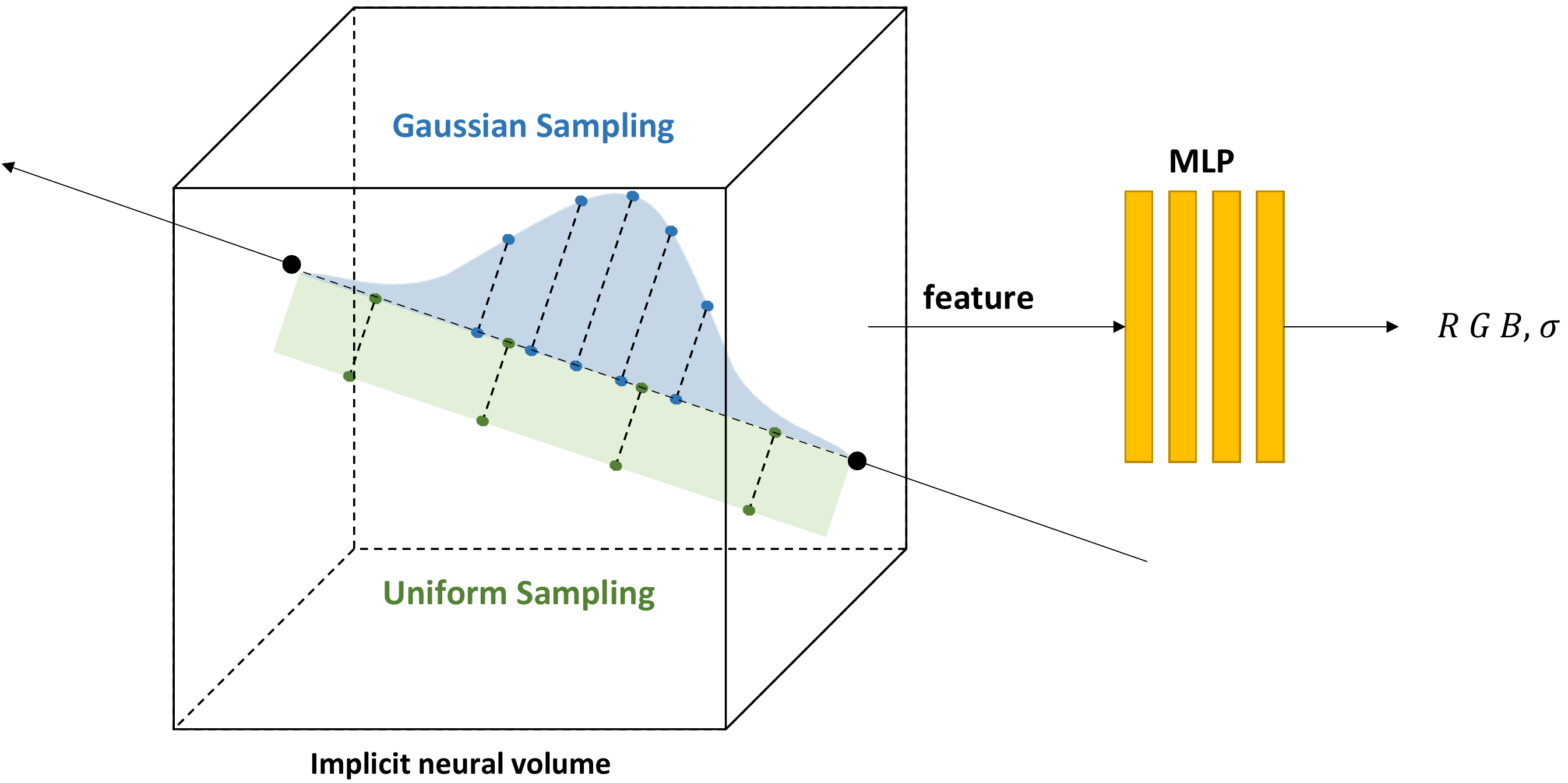}}

    \caption{Demonstration of our proposed Gaussian-Uniform mixture sampling.}
     \label{fig:gaussian}

\end{figure}

To sample points along the ray for volumetric rendering in Eq.~\eqref{eq:render}, existing methods~\cite{chen2021mvsnerf,mildenhall2020nerf,rosu2021neuralmvs,wei2021nerfingmvs,yu2021pixelnerf} partition  $\left[t_{n}, t_{f}\right]$ into $K$ bins following a uniform distribution:
\begin{equation}
 t_{k} \sim \mathcal{U}\left[t_{n}+\frac{k-1}{K}\left(t_{f}-t_{n}\right), t_{n}+\frac{k}{K}\left(t_{f}-t_{n}\right)\right].
\end{equation}
This sampling strategy is inefficient and also independent of depth predictions of the backbone branch. We propose to sample the point candidates following a Gaussian distribution with the guidance of the depth prior from predicted depth map $D$, as illustrated in Fig.~\ref{fig:gaussian}. Assuming the predicted depth for a pixel $\mathbf{p} = (x, y)$ to be $z_\mathbf{p} = D(x, y)$, we sample the candidates using the following distribution:
\begin{equation}
\label{eq:gaussian}
t_{k} \sim \mathcal{N}\left(z_\mathbf{p}, s_\mathbf{p}^{2}\right).
\end{equation}
\begin{equation}
\label{eq:standard}
\text { where } s_\mathbf{p}=\frac{\operatorname{min}\left(|z_\mathbf{p}-t_{f}|,  |z_\mathbf{p}-t_{n}|\right)}{3}.
\end{equation}
Here $z_\mathbf{p}$ and $s_\mathbf{p}$ are the mean and standard deviation of the proposed normal distribution, respectively. This enables geometry features for rendering to be more efficiently optimized, since we sample more point candidates close to the object surface. Furthermore, since we employ differentiable Gaussian sampling, the depth predictions from the backbone branch obtain useful cues on how to improve depth, also in the case of occlusions. In order to make the training stable and help the network to converge, particularly during the beginning stage of the end-to-end training, half of the samples are drawn from the Gaussian distribution from Eq.~\eqref{eq:gaussian} and the other half is uniformly distributed between the near and far planes.

\subsubsection{Depth Rendering Consistency Loss.}
Our rendering consistency network encodes geometry through density estimation. We can convert the density estimate to a depth map in reference view through density integration along the ray:
\begin{equation}
\label{eq:depthrender}
\hat{z}(\mathbf{r})=\sum_{k=1}^{K} w_{k} t_{k}.
\end{equation}
We use the same rendering weights $w_{k}$ as in Eq.~\eqref{eq:render}, and $t_{k}$ is a sampled point candidate as introduced in Sec.~\ref{sec:reference}, following the normal distribution. To enforce that geometry estimated by the backbone branch and by the auxiliary rendering branch match, we minimize the difference between the depth maps with an additional depth rendering consistency loss $L_{DC}$, defined as a Smooth-L1 loss~\cite{girshick2015fast}:
\begin{equation}
\mathcal L_{DC}=\operatorname{smooth}_{L_{1}}\left(\hat{z}(\mathbf{r})-z_\mathbf{p}\right).
\end{equation}
Here, $\hat{z}(\mathbf{r})$ is the integrated depth value from the rendering consistency network and $z_\mathbf{p}$ the corresponding pixel depth value from the backbone network.

\subsection{End-to-end Network Optimization}
\label{sec:networkoptimization}
Unlike pseudo-label-based multi-stage self-supervised methods~\cite{xu2021digging,yang2021self}, our model is trained from scratch in an end-to-end fashion, without any pre-processing or pre-training. The two parts indicated in Sec.~\ref{sec:method}, the backbone and auxiliary network branches, are jointly trained on the whole dataset by minimizing the following overall loss:
\begin{equation}
\begin{array}{r}
\mathcal L=\lambda_{1} \mathcal L_{PC}+\lambda_{2} \mathcal L_{RC}+\lambda_{3}\mathcal L_{DC}\\+\lambda_{4} \mathcal L_{S S I M}+\lambda_{5} \mathcal L_{Smooth}+\lambda_{6} \mathcal L_{DA}.
\end{array}
\end{equation}
Losses $\mathcal L_{S S I M}$ and $\mathcal L_{S m o o t h}$, commonly used in previous works~\cite{dai2019mvs2,huang2021m3vsnet,khot2019learning,xu2021self}, ensure structural similarity and smoothness of predicted depth maps. As proposed by~\cite{xu2021self}, we also apply color fluctuation augmentation to input images and calculate the L1 loss between depth maps predicted under different color augmentations, resulting in $L_{DA}$. We use fixed weights $\lambda_{1}$ = 0.8, $\lambda_{2}$ = 1.0, $\lambda_{3}$ = 1.0, $\lambda_{4}$ = 0.2 and $\lambda_{5}$ = 0.0067. The weight for data augmentation is initialized with $\lambda_{6}=0.01$ and doubled every two epochs. We evaluate the effect of different losses on depth prediction performance in the ablation study (see Sec.~\ref{sec.ablation}).

\section{Experiments}
\label{sec:experiments}
\subsection{Datasets}

The \textbf{DTU} dataset~\cite{aanaes2016large} 
is an indoor dataset with multi-view images and corresponding camera parameters. We follow the settings of MVSNet~\cite{yao2018mvsnet} for dividing training, validation and evaluation sets. There are 124 scenes that are scanned from 49 or 64 views under 7 different light conditions. In the DTU benchmark, models are trained on training set and tested on the evaluation set. We use the official error metrics in DTU to evaluate the \textit{Accuracy}, which is measured as the distance from the result to the structured light reference, indicating the quality of the reconstruction, and \textit{Completeness}, which is measured as the distance from the ground truth reference to the reconstructed result, indicating the proportion of the reconstructed part in the entire point cloud. \textit{Overall} is the average of \textit{Accuracy} and \textit{Completeness} and it reflects the overall quality of the reconstruction.
\textbf{Tanks and Temples}~\cite{knapitsch2017tanks} is a large-scale dataset with various outdoor scenes. It contains an intermediate subset and an advanced subset. The evaluation on this benchmark is conducted online by submitting generated point clouds to the official website. In this benchmark, F-score is calculated for each scene and we compare the mean F-score of intermediate and advanced subset respectively.

\begin{table}[]
  \centering
  \caption{Point cloud evaluation results on DTU~\cite{aanaes2016large}. The lower is better for Accuracy (Acc.), Completeness (Comp.), and Overall. The sections are partitioned into supervised, multi-stage self-supervised and end-to-end (E2E) unsupervised, respectively. The best result is highlighted in bold for each category. All the results other than ours are from previously published literature.}
    \resizebox{0.65\linewidth}{!}
    {\begin{tabular}{lcccc}
    \toprule
        & Method & Acc.$\downarrow$ & Comp.$\downarrow$ & Overall.$\downarrow$\\
    \midrule
     & SurfaceNet \cite{ji2017surfacenet} & 0.450  & 1.040  & 0.745 \\

       &MVSNet \cite{yao2018mvsnet} & 0.396 & 0.527 & 0.462 \\
        &Cas-MVSNet \cite{gu2020cascade} & 0.325 & 0.385 & 0.355  \\
     
    Supervised & PatchmatchNet \cite{wang2021patchmatchnet} & 0.427 & 0.277 & 0.352 \\
    &CVP-MVSNet \cite{yang2020cost} & \textbf{0.296} & 0.406 & 0.351  \\
    &UCSNet \cite{cheng2020deep} & 0.338 & 0.349 & 0.344 \\
    &GBi-Net \cite{mi2021generalized} & 0.315 & \textbf{0.262} & \textbf{0.289}  \\
    \midrule
    Multi-Stage  & Self\_sup CVP-MVSNet \cite{yang2021self} & \textbf{0.308}   & 0.418 & 0.363     \\
    Self-Sup. &U-MVSNet \cite{xu2021digging} & 0.354 & \textbf{0.3535} & \textbf{0.3537} \\
    \midrule

    &Unsup\_MVSNet \cite{khot2019learning} & 0.881 & 1.073 & 0.977  \\
    &MVS2 \cite{dai2019mvs2} & 0.760 & 0.515 & 0.637  \\
E2E UnSup.    &M3VSNet \cite{huang2021m3vsnet} & 0.636 & 0.531 & 0.583 \\
    &JDACS-MS \cite{xu2021self} & 0.398 & 0.318 & 0.358  \\
    &\textbf{RC-MVSNet} & \textbf{0.396} & \textbf{0.295} & \textbf{0.345} \\
    \bottomrule
    \end{tabular}}%
  \label{tab:dtuevaluateresult}%
\end{table}%

\begin{table*}[t]
  \centering
  \caption{Point cloud evaluation results on the intermediate subsets of Tanks and Temples dataset \cite{knapitsch2017tanks}.
  Higher scores are better. The Mean is the average score of all scenes. The sections are partitioned into supervised, multi-stage self-supervised and end-to-end (E2E) unsupervised, respectively. The best result is highlighted in bold for every section.}
    \resizebox{\linewidth}{!}{\begin{tabular}{llccccccccc}
    \toprule
    & & \multicolumn{9}{c}{Tanks\&Temples intermediate} \\
    \midrule
    & Method &  Mean$\uparrow$   & Family$\uparrow$   & Francis$\uparrow$   & Horse$\uparrow$   & Lighthouse$\uparrow$   & M60$\uparrow$    & Panther$\uparrow$   & Playground$\uparrow$   & Train$\uparrow$  \\
    \midrule
    &MVSNet~\cite{yao2018mvsnet}  & 43.48 & 55.99 & 28.55 & 25.07 & 50.79 & 53.96 & 50.86 & 47.90  & 34.69 \\
    &CIDER~\cite{xu2020learning}&  46.76 & 56.79 & 32.39 & 29.89 & 54.67 & 53.46 & 53.51 & 50.48 & 42.85\\
    Supervised&PatchmatchNet~\cite{wang2021patchmatchnet} & 53.15 & 66.99 & 52.64 & 43.24 & 54.87 & 52.87 & 49.54 & 54.21 & \textbf{50.81} \\
    &CVP-MVSNet~\cite{yang2020cost} & 54.03& \textbf{76.50}& 47.74& 36.34 &55.12& \textbf{57.28}& \textbf{54.28}& 57.43 &47.54
 \\

    &UCSNet~\cite{cheng2020deep}& 54.83& 76.09& 53.16& 43.03 &54.00& 55.60& 51.49& 57.38 &47.89\\ 
        &Cas-MVSNet~\cite{gu2020cascade} & \textbf{56.42} & 76.36 & \textbf{58.45}& \textbf{46.20} &\textbf{55.53} &56.11 &54.02 &\textbf{58.17} &46.56 \\
    \midrule
    
    Multi-Stage&Self\_sup CVP-MVSNet \cite{yang2021self}&46.71 & 64.95 &38.79 &24.98 & 49.73  & 52.57 &   \textbf{51.53} & 50.66&  40.45\\

   Self-Sup.& U-MVSNet~\cite{xu2021digging}&\textbf{57.15} &\textbf{76.49} &\textbf{60.04} &\textbf{49.20} &\textbf{55.52} &\textbf{55.33} &51.22 &\textbf{56.77} &\textbf{52.63} \\

    \midrule
    &MVS2~\cite{dai2019mvs2}  & 37.21 &47.74 &21.55 &19.50 &44.54 &44.86 &46.32 &43.38 &29.72 \\
   E2E & M3VSNet ~\cite{huang2021m3vsnet}  & 37.67 &47.74 &24.38 &18.74 &44.42 &43.45 &44.95 &47.39  &30.31 \\
  Unsup. & JDACS-MS~\cite{xu2021self} & 45.48 &66.62 &38.25 &36.11 &46.12 &46.66 &45.25 &47.69 &37.16\\
   & \textbf{RC-MVSNet} & {\textbf{55.04}} & \textbf{75.26} & \textbf{53.50} &\textbf{45.52} & \textbf{53.49} & \textbf{54.85} & \textbf{52.30} & \textbf{56.06} & \textbf{49.37} \\
    \bottomrule
    \end{tabular}}%
  \label{tab:tanksanvancedintermediate}%
\end{table*}%

\subsection{Implementation Details} \label{sec:ImplementationDetails}
\noindent{\textbf{Training Details.}} 
The proposed RC-MVSNet is trained on the DTU dataset. Following~\cite{dai2019mvs2,huang2021m3vsnet,khot2019learning,xu2021self}, we use the high-resolution DTU data provided by the open source code of MVSNet \cite{yao2018mvsnet}. We first resize the input images to $600 \times 800$, following previous methods. Then we crop resized images into $512 \times 640$ patches. We adopt the backbone of Cas-MVSNet~\cite{gu2020cascade} to construct our multi-scale pipeline with 3 stages. For each stage, we use different feature maps and the 3D-CNN network parameters. The whole network is optimized by an Adam optimizer in Pytorch for 15 epochs with an initial learning rate of 0.0001, which is down-scaled by a factor of 2 after 10, 12, and 14 epochs. We train with a batch size of 4 using four NVIDIA RTX 3090 GPUs.

\begin{figure}
  \centering
  \includegraphics[width=\textwidth]{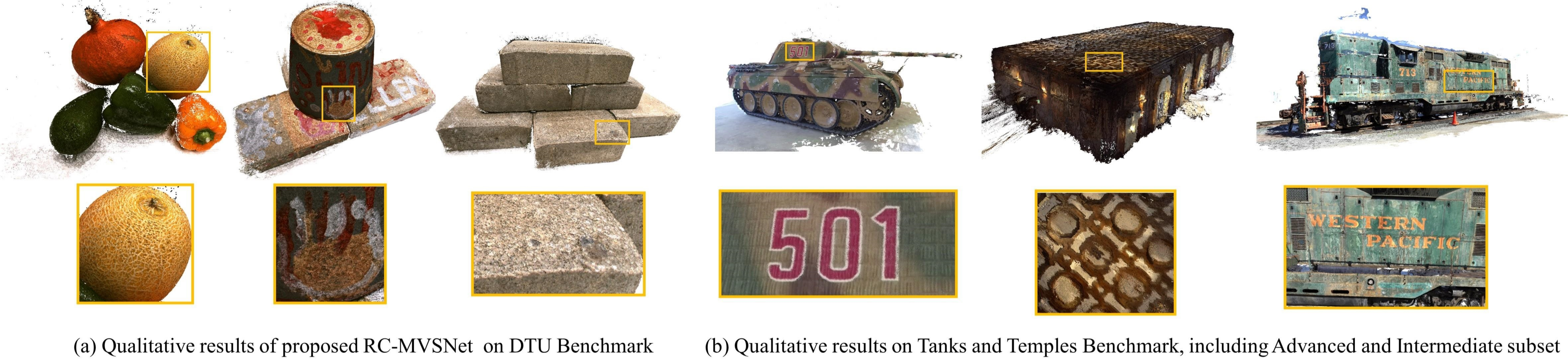}
  \caption{Point cloud visualization of our method on DTU~\cite{aanaes2016large} and Tanks and Temples~\cite{knapitsch2017tanks}.}
     \label{fig:dtuandtanks}
\end{figure}

\noindent{\textbf{Testing Details.}}
The model trained on DTU training set is used for testing on DTU testing set. The input image number $N$ is set to 5, each with a resolution of $1152 \times 1600$. It takes 0.475 seconds to test each sample. The model trained on DTU training dataset is directly used for testing on Tanks and Temples intermediate and advanced datasets without finetuning. The image sizes are set to $1024 \times 1920$ or $1024 \times 2048$ and the input image number $N$ is set to 7. We then filter predicted depth maps of a scene using photometric and geometric consistencies and fuse then into one point cloud. We define geometric and photometric consistencies similarly to the ones used in MVSNet \cite{yao2018mvsnet}, a detailed description can be found in supplemental material. Fig.~\ref{fig:dtupoint} shows visualizations of point clouds of RC-MVSNet and previous unsupervised methods, and Fig.~\ref{fig:dtuandtanks} shows  reconstructions of our method on the test dataset.

\subsection{Benchmark Performance}
\par\noindent{\textbf{Evaluation on DTU Dataset.}} 
We evaluate the depth prediction performance on the DTU test set, and compare with previous state-of-the-art methods in Table~\ref{tab:dtuevaluateresult}. 
Our RC-MVSNet architecture achieves the best accuracy, completeness and overall score (lower is better for all metrics) among all end-to-end unsupervised methods.
Our model improves the overall score from \textbf{0.358} of JDACS~\cite{xu2021self} to \textbf{0.345}. The completeness is improved by \textbf{0.023}, demonstrating the effectiveness of our method to resolve incomplete predictions caused by occlusions. The overall score is also better than the multi-stage self-supervised approaches, and even most supervised methods. In addition to the performance of point cloud reconstruction, we also provide evaluations of predicted depth maps. We compare them with supervised MVSNet~\cite{yao2018mvsnet} and previous unsupervised approaches~\cite{dai2019mvs2,huang2021m3vsnet,khot2019learning,xu2021self}, the results are summarized in Table.~\ref{tab:thresratiocompare}. Our method again achieves state-of-the-art accuracy. Visualizations of our depth predictions can be found in the supplementary material.

\par\noindent{\textbf{Evaluation on Tanks and Temples.}} 
We train the proposed RC-MVSNet on the DTU train set, and test on the Tanks and Temples dataset without finetuning. We compare our method to state-of-the-art supervised, pseudo-label-based multi-stage self-supervised methods and end-to-end unsupervised methods. Table~\ref{tab:tanksanvancedintermediate} shows the evaluation results on the intermediate subset. Our RC-MVSNet achieves the best performance among unsupervised methods. In particular, our mean score achieves significant improvement of \textbf{+17.37} over M3VSNet~\cite{huang2021m3vsnet} and \textbf{+9.56} over JDACS~\cite{xu2021self}. Moreover, we also obtain state-of-the-art performance in all sub-scenes, which fully confirms the effectiveness of our method. Our \textbf{anonymous} evaluation on the leaderboard~\cite{tanksandtemplesleader} is named as RC-MVSNet. The results of advanced subset can be found in the supplementary material, along with additional qualitative results and discussion. 
\begin{table}[t]
  \centering
  \caption{Depth map evaluation results in terms of accuracy on DTU evaluation set~\cite{aanaes2016large} (higher is better). All thresholds are given in millimeters.}
  {\begin{tabular}{cccc}
  \toprule
    Method & $<$2$\uparrow$ & $<$4$\uparrow$ & $<$8$\uparrow$ \\
    \midrule
    MVSNet \cite{yao2018mvsnet}  & 0.704 & 0.778 & 0.815\\
    MVSNeRF \cite{wang2021patchmatchnet}  & 0.510 & 0.645 & 0.734  \\
    Unsup\_MVSNet \cite{khot2019learning}   & 0.317 & 0.384 & 0.402 \\
    M3VSNet \cite{huang2021m3vsnet}   & 0.603 & 0.769 & 0.857  \\
    JDACS-MS \cite{xu2021self}   & 0.553 & 0.705 & 0.786   \\

    \midrule
    \midrule
    \textbf{RC-MVSNet}  & \textbf{0.730} & \textbf{0.795}  & \textbf{0.863} \\
    \bottomrule
    \end{tabular}}%
    \label{tab:thresratiocompare}%
\end{table}%

\subsection{Ablation Study} 
\label{sec.ablation}
We provide an ablation analysis, demonstrating the point cloud reconstruction performance gain of each component of our proposed method. We also evaluate performance when sampling different number of rays during rendering in the supplementary material.
\begin{figure}
  \centering
    \subfigure[$L_{PC}$+$L_{DA}$]{              
        \includegraphics[width=0.22\textwidth]{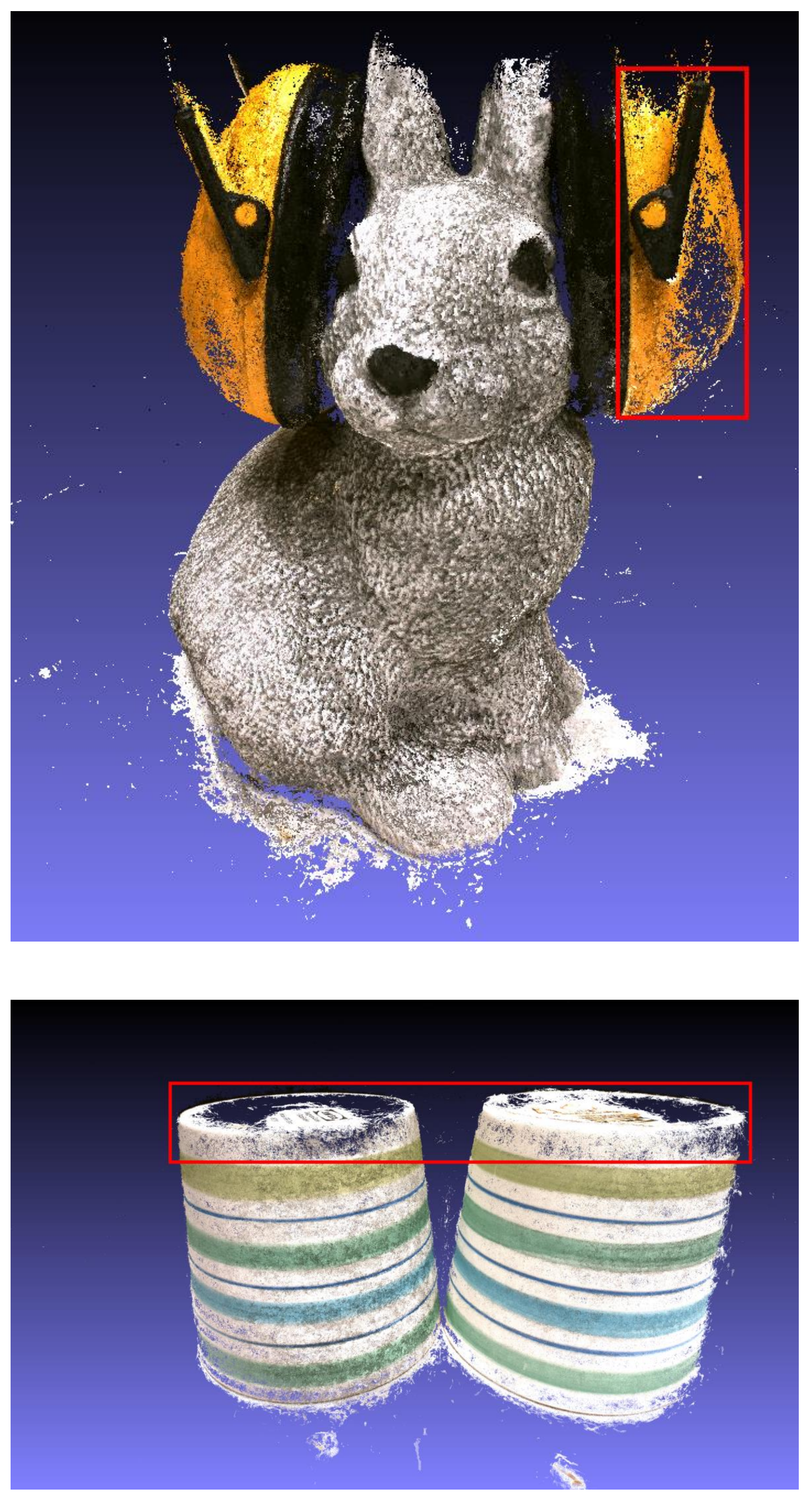}}
    \subfigure[$L_{PC}$+$L_{DA}$\newline+$L_{RC}$]{
        \includegraphics[width=0.22\textwidth]{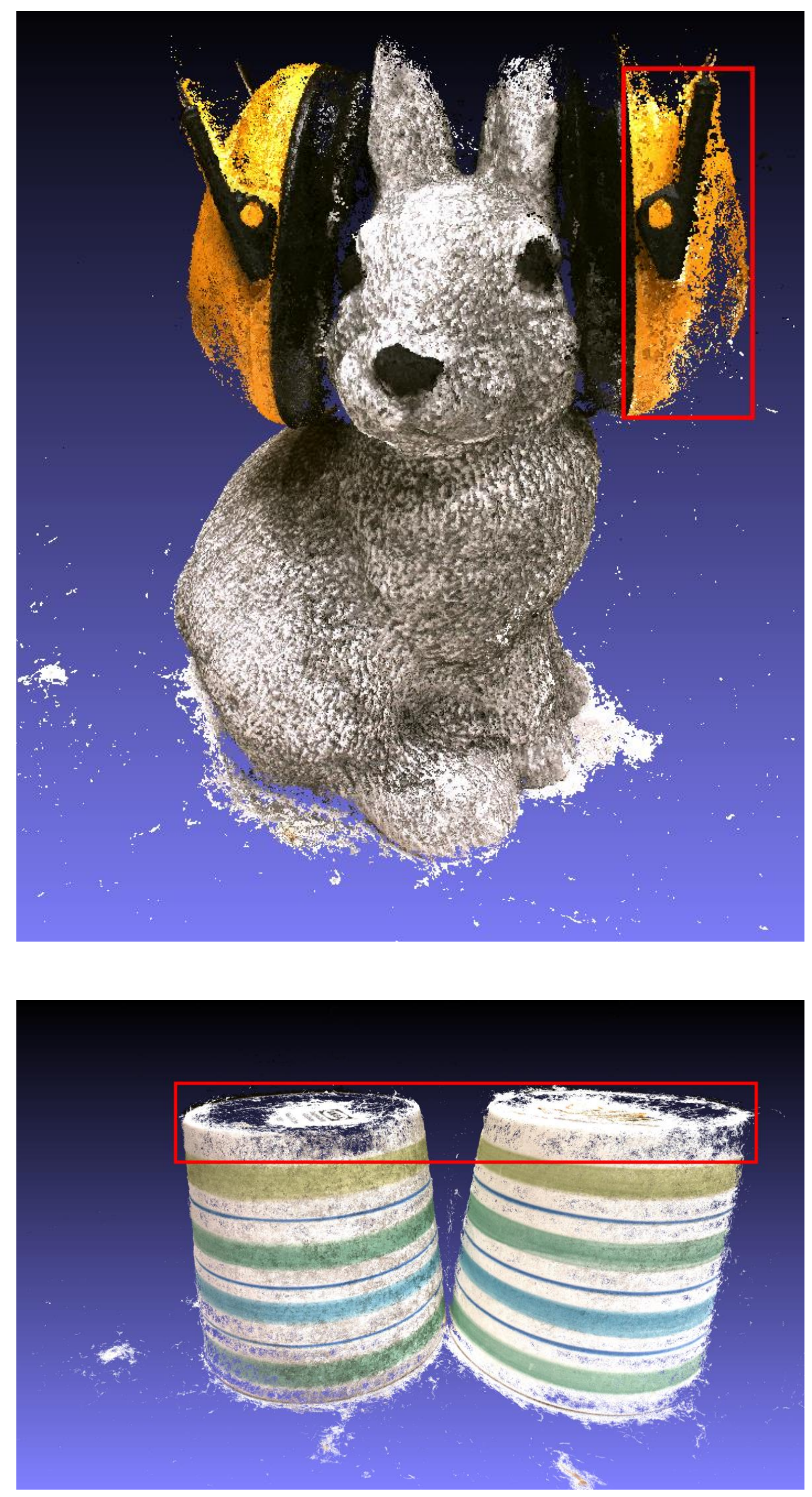}}
    \subfigure[$L_{PC}$+$L_{DA}$\newline+$L_{RC}$+$L_{DC}$+G-U]{              
        \includegraphics[width=0.22\textwidth]{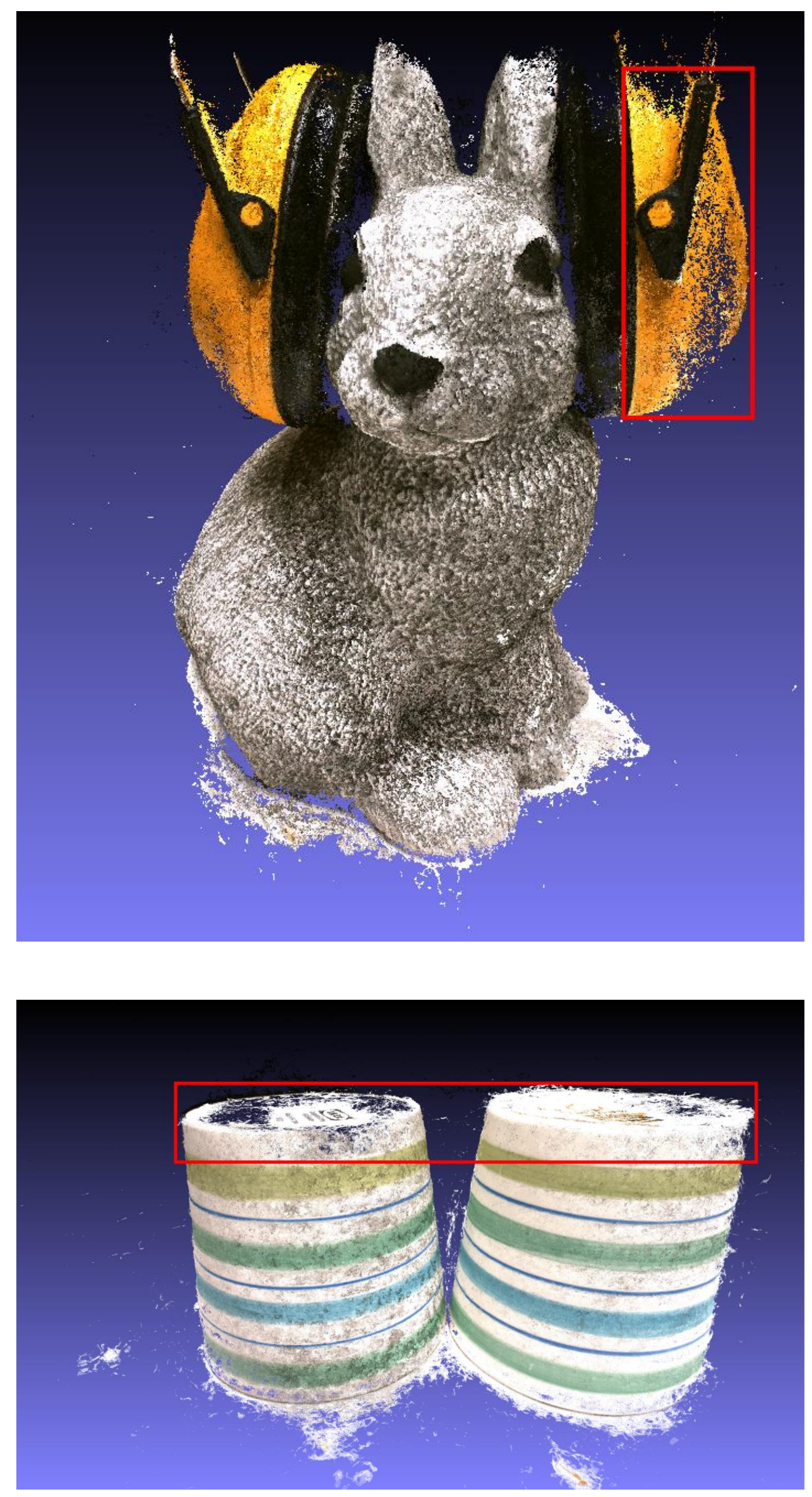}}
    \subfigure[GT]{              
        \includegraphics[width=0.22\textwidth]{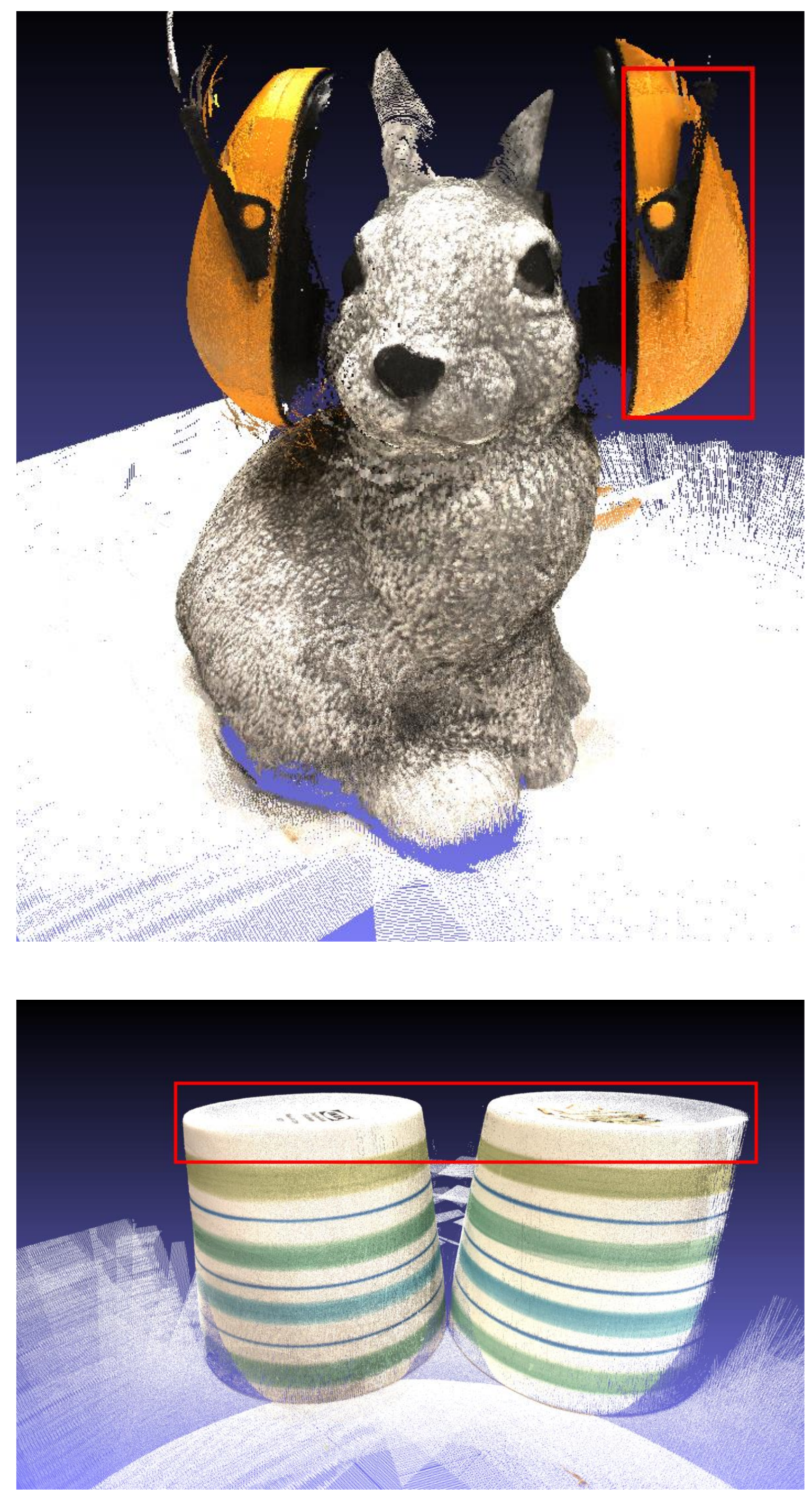}}
    \caption{Qualitative results of RC-MVSNet on scan33 (top) and scan48 (bottom) of the DTU dataset. 
    }
     \label{fig:ablation}
\end{figure}

\subsubsection{Effect of each component of the unsupervised approach.}
To evaluate the performance gain of each part in our method, we provide qualitative comparison of models trained using different combinations of core contributions in Fig.~\ref{fig:ablation}. We can easily observer that both reference view synthesis and depth rendering consistency yield better reconstruction results, especially when considering completeness. Quantitative results are listed in Table.~\ref{tab:component}, further confirming the importance of all proposed network components.
\begin{table}[htbp]
  \centering
  \caption{Ablation study of different components of our proposed unsupervised approach (the lower is better). $L_{PC}$: Photometric Consistency Loss. $L_{DA}$: Data Augmentation Loss.  $L_{RC}$: Reference View Synthesis Loss. G-U: Gaussian-Uniform Mixture Sampling. $L_{DC}$: Depth Rendering Consistency Loss. The last three are contributions in this work.}
   \resizebox{0.6\linewidth}{!}{
    \begin{tabular}{ccccc|ccc}
    \toprule
    \multicolumn{1}{l}{$L_{PC}$} &\multicolumn{1}{l}{$L_{DA}$} & \multicolumn{1}{l}{$L_{RC}$} &\multicolumn{1}{l}{G-U} & \multicolumn{1}{l|}{$L_{DC}$} & \multicolumn{1}{l}{Acc$\downarrow$} & \multicolumn{1}{l}{Comp$\downarrow$} & \multicolumn{1}{l}{Overall$\downarrow$} \\
    \midrule
    \midrule
   \checkmark  &    &      & &       & 0.462	 & 0.328 & 	0.395 \\
    \midrule   
   \checkmark &  \checkmark    &  &     &       & 0.419& 	0.346&	0.383
 \\
    \midrule
   \checkmark & \checkmark    & \checkmark   &   &       & 0.415 & 0.300 & 0.357 \\
    \midrule
      \checkmark & \checkmark    & \checkmark    &  \checkmark&       & 0.404 & 0.299 & 0.352 \\
    \midrule
    \checkmark  &   \checkmark  &   \checkmark    &  &   \checkmark     & 0.397	 & 0.299 & 0.348	\\
    \midrule
   \checkmark & \checkmark      & \checkmark   &  \checkmark& \checkmark      & \textbf{0.396}   & \textbf{0.295} & \textbf{0.345} \\
    \bottomrule
    \end{tabular}}%
  \label{tab:component}%
\end{table}%

\section{Conclusions}
We show that unsupervised multi-view-stereo can be improved by challenging the previous assumption of photometric consistency. We thus propose a novel unsupervised MVS approach based on rendering consistency (RC-MVSNet). To handle the ambiguous supervision, we propose a reference view synthesis loss via differentiable volumetric rendering. To solve the incompleteness caused by occlusions, we introduce Gaussian-Uniform mixture sampling to learn geometry features close to the object surface. To further improve the robustness and smoothness of the depth map, we propose a depth rendering consistency loss. The experiments demonstrate the effectiveness of our RC-MVSNet approach. 

\section{Acknowledgments}
This project is funded by the ERC Starting Grant Scan2CAD (804724), a TUM-IAS Rudolf Mößbauer Fellowship, and the German Research Foundation (DFG) Grant Making Machine Learning on Static and Dynamic 3D Data Practical.

\clearpage
\bibliographystyle{splncs04}
\bibliography{main}
\newpage

\par \noindent \textbf{\Large{Supplementary Material}}\\
\par \indent In this supplementary material, we present our results on the Tanks\&Temples Advanced subset in Sec.~\ref{sec:advance} and describe depth map fusion in detail in Sec.~\ref{sec:fusion}. We provide details about cost volume regularization of the Cas-MVSNet backbone and implicit volume construction of our rendering consistency network in Sec.~\ref{sec:cost}. We also show additional depth map prediction and point cloud reconstruction results in Sec.~\ref{sec:visualization}. Furthermore, we provide an ablation study on number of sampled rendering rays in Sec.~\ref{sec:additionalablation}. Finally, we discuss training strategy of pseudo label based multi-stage self-training and end-to-end unsupervised training in Sec.~\ref{sec:discussion}.

\section{Performance on Tanks\&Temples Advanced Benchmark}
\label{sec:advance}
We train the proposed RC-MVSNet on the DTU training set, and test on the Tanks\&Temples Advanced dataset without finetuning. We compare our method to state-of-the-art supervised~\cite{gu2020cascade,wang2021patchmatchnet,xu2020learning,yao2019recurrent}, pseudo-label-based multi-stage self-supervised method U-MVSNet~\cite{xu2021digging}. Table~\ref{tab:tanksanvanced} shows the evaluation results on the advanced subset. Our RC-MVSNet is the first end-to-end unsupervised method on advanced subset of Tanks\&Temples. We achieve comparable performance to the multi-stage self-supervised method~\cite{xu2021digging} and supervised methods~\cite{gu2020cascade,wang2021patchmatchnet}. We also outperform supervised method CIDER~\cite{xu2020learning} by \textbf{+7.7 (33\%)} and R-MVSNet~\cite{yao2019recurrent} by \textbf{+5.91 (24\%)}.

\section{Depth Map Fusion}
\label{sec:fusion}

After obtaining depth maps through RC-MVSNet, we need to convert them into 3D point clouds through depth map fusion for further evaluation. 
Following MVSNet \cite{yao2018mvsnet} and R-MVSNet \cite{yao2019recurrent}, we use geometric consistency and photometric consistency to remove occlusions and unreliable regions. 
Geometric consistency projects depth maps of source images into the reference image and masks out depth-inconsistency regions. To unify point cloud representation, we compute the average depth value of the consistent region.
As for photometric consistency, we directly use the confidence maps generated by RC-MVSNet and only keep the regions with high confidence.
Finally, we re-project the pixels that satisfy geometric consistency and photometric consistency to the world coordinate system to generate point clouds.

\begin{figure}
  \centering
    \includegraphics[width=\textwidth]{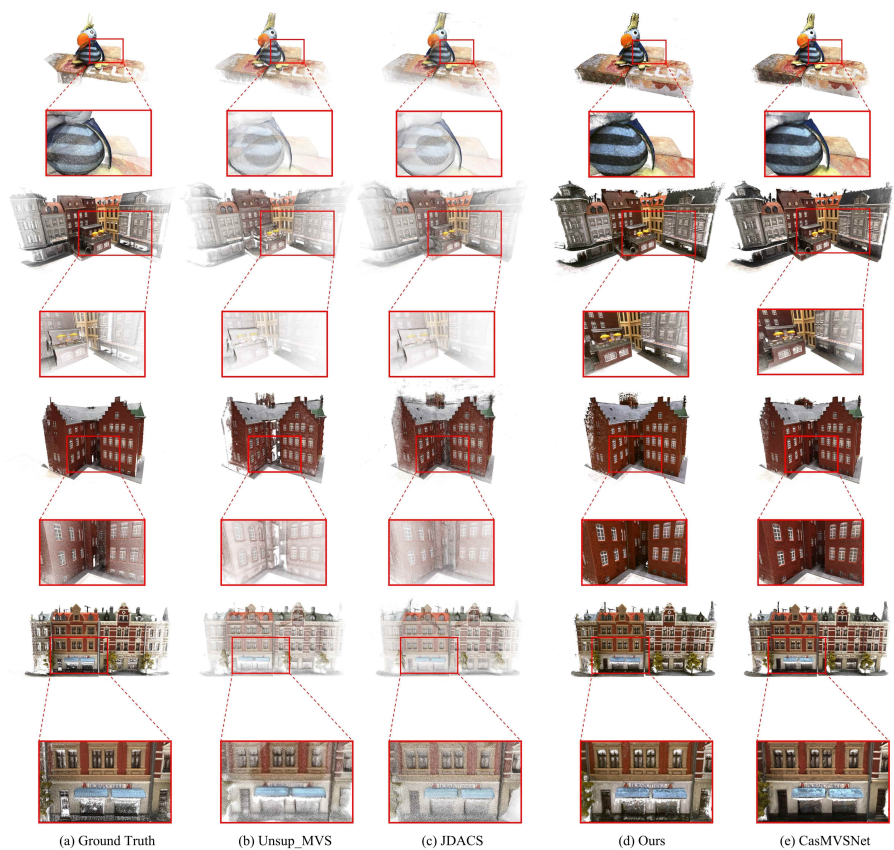}
    \caption{Qualitative comparison of point cloud reconstructions on DTU.}
     \label{fig:dtu}
\end{figure}

\begin{figure}
  \centering
    \includegraphics[width=\textwidth]{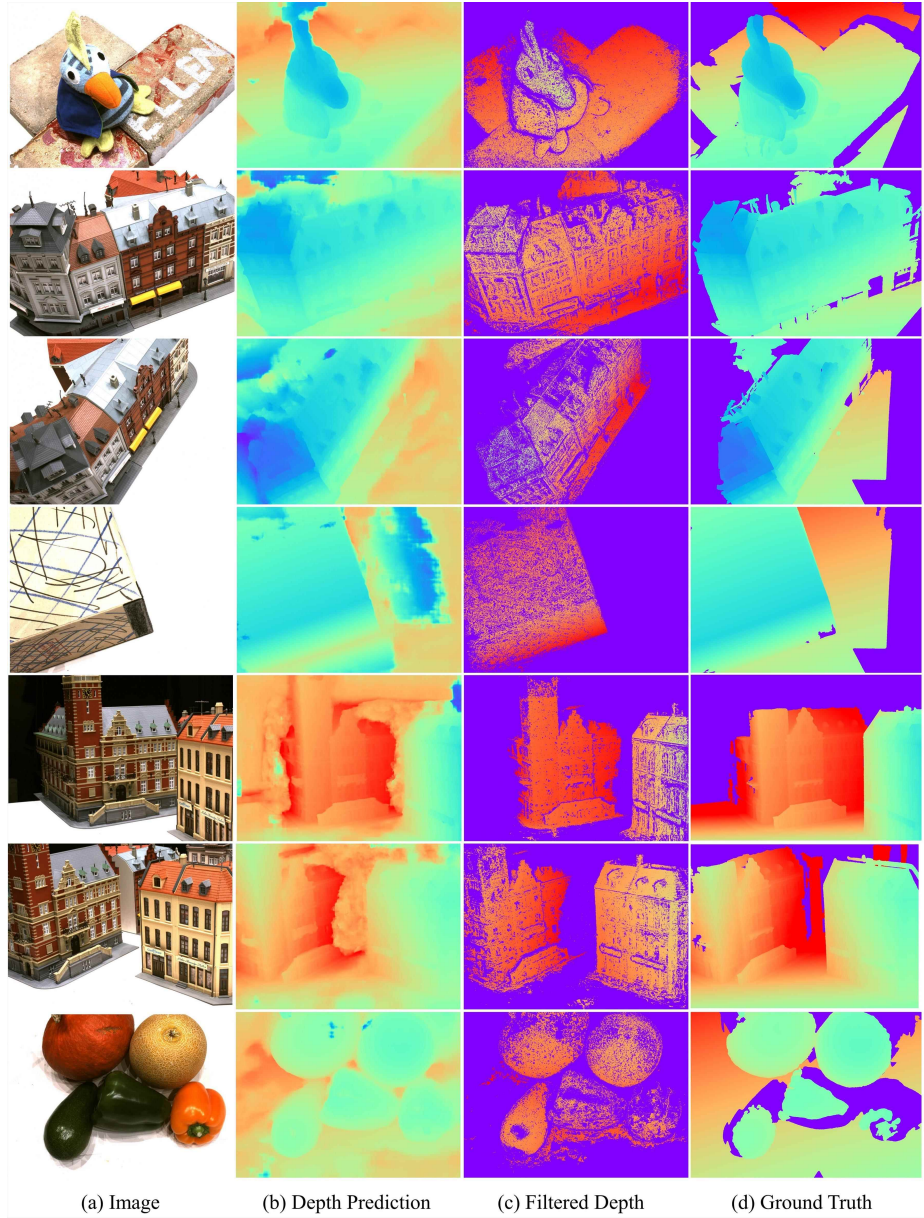}
    \caption{Visualization of the prediction and filtered depth maps.}
     \label{fig:depthmaps}
\end{figure}
\begin{figure}
  \centering
    \includegraphics[width=\textwidth]{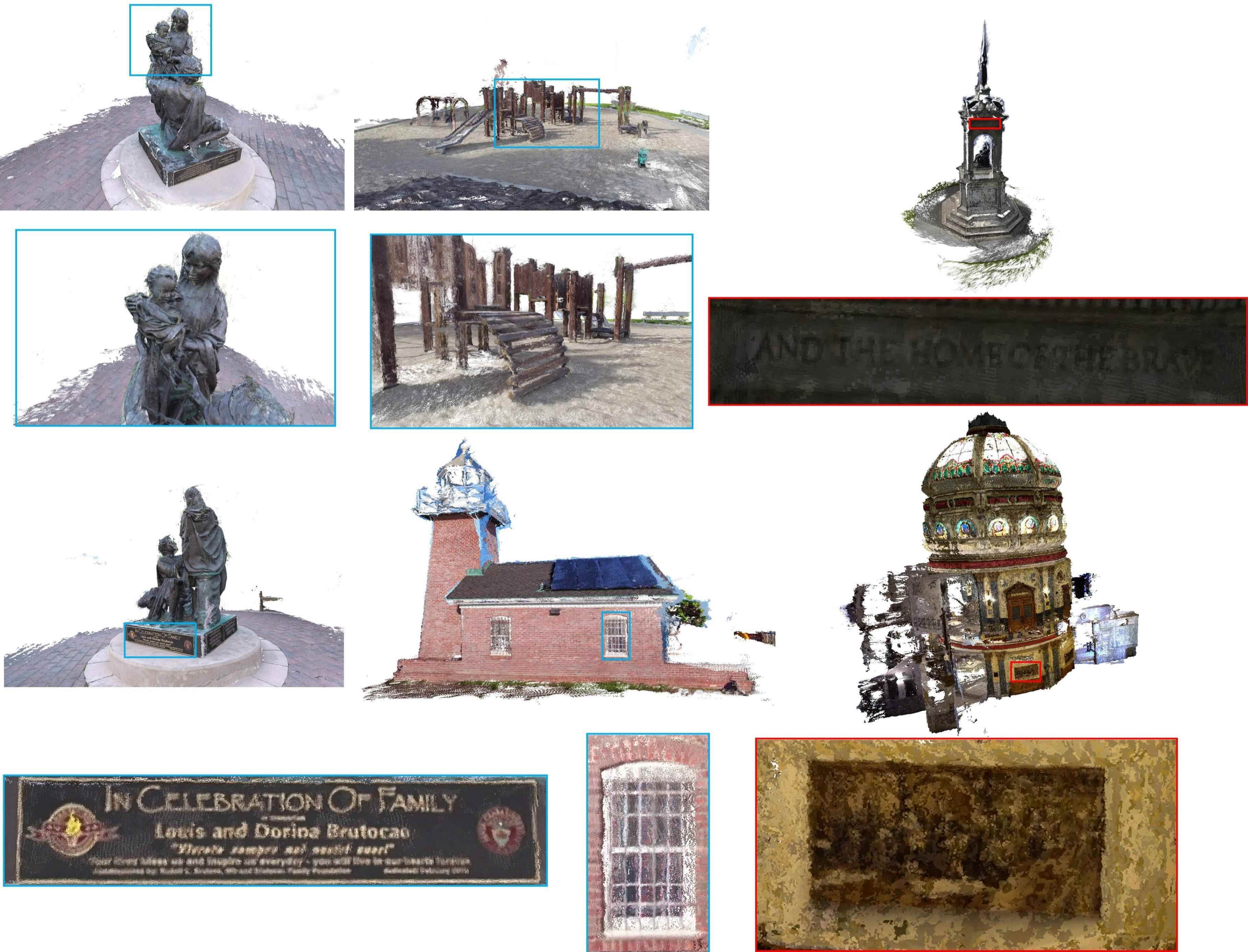}
    \caption{Qualitative results of point cloud reconstructed on Tanks\&Temples.}
     \label{fig:tanks}
\end{figure}
\section{Cost Volume and Implicit Neural Volume}
\label{sec:cost}
\begin{table*}[t]
  \centering
  \caption{Point cloud evaluation results on the advanced subset of Tanks\&Temples dataset \cite{knapitsch2017tanks}.
  Higher scores are better. The Mean is the average score of all scenes. The sections are partitioned into supervised, multi-stage self-supervised and end-to-end unsupervised, respectively. The best result is highlighted in bold for each section.}
    \resizebox{\linewidth}{!}{\begin{tabular}{llccccccc}
    \toprule
    & & \multicolumn{7}{c}{Tanks\&Temples advanced}  \\
    \midrule
    & Method & Mean$\uparrow$   & Auditorium$\uparrow$   & Ballroom$\uparrow$   & Courtroom$\uparrow$   & Museum$\uparrow$   & Palace$\uparrow$   & Temple$\uparrow$ \\
    \midrule
    & CIDER~\cite{xu2020learning} &  23.12&  12.77&  24.94 & 25.01 & 33.64 & 19.18 & 23.15  \\
    Supervised &R-MVSNet~\cite{yao2019recurrent} & 24.91 &12.55& 29.09& 25.06& 38.68& 19.14 &24.96 \\
    &CasMVSNet~\cite{gu2020cascade} & 31.12 & 19.81	& \textbf{38.46} & 29.10	&\textbf{43.87}	&27.36	&28.11  \\
    &PatchmatchNet~\cite{wang2021patchmatchnet} & \textbf{32.31} &\textbf{23.69} & 37.73 & \textbf{30.04} & 41.80 & \textbf{28.31} & \textbf{32.29}  \\
    \midrule

     Multi-Stage Self-sup.&U-MVSNet~\cite{xu2021digging} & \textbf{30.97} & \textbf{22.79} & \textbf{35.39} &\textbf{28.90} &\textbf{36.70} &\textbf{28.77} &\textbf{33.25} \\

    \midrule
   
    E2E Unsup.&\textbf{RC-MVSNet} & \textbf{30.82} & \textbf{21.72} & \textbf{37.22} & \textbf{28.62}  & \textbf{37.37} & \textbf{27.88} & \textbf{32.09} \\
    \bottomrule
    \end{tabular}}%
  \label{tab:tanksanvanced}%
\end{table*}%
Arbitrary fully-supervised MVS network could be used as our backbone in our framework -- we use CasMVSNet~\cite{gu2020cascade} as default backbone. The 2D CNN extract latent features from $N$ input views. Then the features from the $N -1$ source views are reprojected into the reference view via differential homography warping, as following: 
\begin{equation}
H_{j}(d)=K_{j} \cdot R_{j} \cdot\left(I-\frac{\left(t_{1}-t_{j}\right) \cdot n_{1}^{T}}{d}\right) \cdot R_{1}^{T} \cdot K_{1}^{-1}
\end{equation}

\par \noindent where $H_{j}(d)$ denote the homography between the feature
maps of the $j^{th}$($2 \leq j \leq N$) view and the reference feature map at depth $d$. The camera intrinsics $K_{j}$, rotations $R_{j}$ and $t_{j}$  are also given according to $j^{th}$ view respectively. $n_{1}$ refers to the principle axis of the reference camera. The variance of these feature maps are calculated to construct a cost volume, which is regularized by 3D CNNs in each stage of the cascade structure. 
After the 3D convolutions, a pixel-wise depth map is regressed with soft-argmax upon the depth dimension of the probability volume.

 For the reference volume ${{V}_{1}}$ of reference image $I_{1}$ and warped volume $\left\{{V}_{j}\right\}_{j=2}^{N}$ of source images $\left\{{I}_{j}\right\}_{j=2}^{N}$, the cost volume $C$ in the backbone for depth estimation was constructed by Eq. 1. in our paper. This variance volume contains image appearance information and camera poses across all input views. However this volume is used for geometry reconstruction of depth inference, specifically for the reference view. We expect to use the information from \textit{only} source views to synthesize the reference view. Hence, we calculate the variance volume ${C^{\prime}}$ from warped volume $\left\{{V}_{j}\right\}_{j=2}^{N}$ by:
\begin{equation}
{C^{\prime}}={Var}\left({V}_{2}, \cdots, {V}_{N}\right)=\frac{\sum_{j=2}^{N}\left({V}_{j}-\overline{{V}}_{j}^{\prime}\right)^{2}}{N-1}
\end{equation}
where $Var$ denotes the same calculation and $\overline{{V}}_{j}^{\prime}$ is the mean of warped volumes. In this way, we aggregate the information across $N-1$ source views construct the implicit neural volume.

\section{Additional Qualitative Results}
\label{sec:visualization}

\subsection{Depth Map Visualization on DTU Benchmark}
Fig.~\ref{fig:depthmaps} provides visualization of depth map of scans 4, 9, 10, 29 and 75 of DTU benchmark~\cite{aanaes2016large}.

\subsection{Point Cloud Visualization on DTU Benchmark}
Fig.~\ref{fig:dtu} provides additional reconstruction visualization on DTU benchmark\cite{aanaes2016large}. Our unsupervised model shows significant improvement compared to previous state-of-the-arts, and achieves comparable reconstruction results to the supervised approach Cas-MVSNet\cite{gu2020cascade}.

\subsection{Point Cloud Visualization on Tanks\&Temples Benchmark}
Fig.~\ref{fig:tanks} visualizes additional point cloud reconstruction results on Tanks\&Temples benchmark. Our method produces accurate and complete reconstructions.

\section{Additional Ablation Study}
\label{sec:additionalablation}
\subsubsection{Number of sampled rays for reference view synthesis}
Due to limited memory usage, we're not able to render complete depth maps and images during training times. Following common setting~\cite{chen2021mvsnerf,mildenhall2020nerf}, we only sampled a subset of rays during the volumetric rendering process. The performance of using different number of sampled rays is shown in Table.~\ref{tab:numrays}. 
\begin{table}[t]
  \centering
  \caption{Performance at different number of sampled rays during volumetric rendering}
    \begin{tabular}{c|ccc|c|c|c|c}
    \toprule
    \multicolumn{1}{c|}{Num\_rays} & \multicolumn{1}{c}{Acc$\downarrow$} & \multicolumn{1}{c}{Comp$\downarrow$} & \multicolumn{1}{c}{Overall$\downarrow$} & \multicolumn{1}{|c}{Train Mem}& \multicolumn{1}{|c}{Train Img Size} & \multicolumn{1}{|c}{Test Mem} &\multicolumn{1}{|c}{Test Img Size}\\
    \midrule
    \midrule    
    256  & 0.404  & 0.296 & 0.350 & 14.6 GB & $640\times512$ & 7.5 GB&$1600\times1152$  \\
    \midrule
    1024  &0.396 &\textbf{0.295 } & \textbf{0.345} & 15.5 GB & $640\times512$ & 7.5 GB &$1600\times1152$\\
    \midrule
    4096  & 0.400 & 0.299 & 0.350 & 18.3 GB & $640\times512$ & 7.5 GB & $1600\times1152$  \\
    \midrule
    8192  & \textbf{0.395} & 0.300 & 0.348 & 23.3 GB & $640\times512$ & 7.5 GB & $1600\times1152$ \\
    \bottomrule
    \end{tabular}%
  \label{tab:numrays}%
\end{table}%

\section{Discussion}
\label{sec:discussion}
As we described in the paper, multi-stage self-supervised methods suffer from complicated pre-training and pre-processing. The limitation of training time makes it difficult for these methods to be applied in practical scenarios. According to U-MVSNet~\cite{xu2021digging}, the pretraining of PWC-Net on DTU\cite{aanaes2016large} and whole self-supervision training stage take 16 epochs in total. Then the post-training based on generated pseudo label takes further 16 epochs. We use the same backbone as them and it only takes 15 epochs to converge with 6 hours per epoch on NVIDIA RTX 3090. For self-supervised CVP-MVSNet~\cite{yang2021self}, the self-training takes 15 hours per epoch on NVIDIA RTX 2080Ti. Hence, improving the efficiency of previous learning-based methods, both running time and memory usage, while maintaining comparable performance with self-supervised and supervised ones, can be regarded as one of our method's advantages.

\end{document}